\documentclass[letterpaper, 10 pt, conference]{ieeeconf}
\IEEEoverridecommandlockouts
\usepackage[utf8]{inputenc}
\usepackage[T1]{fontenc}
\usepackage{import}
\usepackage{bbm}
\usepackage{graphicx}
\usepackage{makecell}
\usepackage[export]{adjustbox}
\usepackage[cmex10]{amsmath}
\usepackage{amssymb}
\usepackage{amsfonts}
\usepackage{textcomp}
\usepackage{booktabs}
\usepackage{multirow}
\usepackage{url}

\usepackage{amsthm}
\usepackage{amsmath}
\usepackage{cite}
\usepackage[caption=false,font=footnotesize]{subfig}
\usepackage[usenames, dvipsnames]{color}
\usepackage{overpic}
\usepackage[table]{xcolor}
\usepackage[section]{placeins}
\usepackage{graphicx}
\usepackage{complexity}
\usepackage{balance}
\usepackage{siunitx}
\usepackage[normalem]{ulem}
\usepackage{siunitx}
\usepackage[colorlinks=true,allcolors=black,breaklinks]{hyperref}

\newcommand{\todo}[1]{}

\title{Estimating Spatially-Dependent GPS Errors Using a Swarm of Robots}
\author{Praneeth Somisetty, Robert Griffin, Victor M. Baez, Miguel F. Arevalo-Castiblanco,\\{} Aaron T. Becker, Jason M. O'Kane%
\thanks{P. Somisetty and J. M. O'Kane are with Texas A\&M University, College Station, Texas, USA.  R. Griffin, V. Baez, M. F. Arevalo-Castiblanco, and A. T. Becker are with the University of Houston.
$\{ \texttt{praneeth.1002@gmail.com}, \texttt{jokane@tamu.edu} \} \cup \{ \texttt{rrgriffi}, \texttt{vjmontan}, \texttt{farevalo2}, \texttt{atbecker} \} \times \{ \texttt{@uh.edu} \}$}
\thanks{Research was sponsored in part by the DEVCOM Analysis Center, accomplished under Contract Number W911QX-23-D-0009 and the Army Research Laboratory,  accomplished under Cooperative Agreement Number W911NF-23-2-0014. The views and conclusions contained in this document are those of the authors and should not be interpreted as representing the official policies, either expressed or implied, of the DEVCOM Analysis Center, the Army Research Laboratory, or the U.S. Government. The U.S. Government is authorized to reproduce and distribute reprints for Government purposes notwithstanding any copyright notation herein.
}}

\begin{document}

\maketitle

\begin{abstract}
External factors, including urban canyons and adversarial interference, can lead to Global Positioning System (GPS) inaccuracies that vary as a function of the position in the environment. This study addresses the challenge of estimating a static, spatially-varying error function using a team of robots. We introduce a State Bias Estimation (SBE) algorithm whose purpose is to estimate the GPS biases.
The central idea is to use sensed estimates of the range and bearing to the other robots in the team to estimate changes in bias across the environment.
A set of drones moves in a 2D environment, each sampling data from GPS, range, and bearing sensors. 
The biases calculated by the SBE at estimated positions are used to train a Gaussian Process Regression (GPR) model.
We use a sparse Gaussian process–based Informative Path Planning (IPP) algorithm that identifies high-value regions of the environment for data collection. The swarm plans paths that maximize information gain in each iteration, further refining their understanding of the environment’s positional bias landscape.
%
We evaluated SBE and IPP in simulation and compared the IPP methodology to an open-loop strategy. 
\todo{The algorithms estimate error over a $\SI{50}{\meter}\times \SI{50}{\meter}\times \SI{3}{\meter}$, with GPS biases ranging from 0 to $\SI{10}{\meter}$, with an RMS error of ??, after a run time of $\SI{20}{\second}$ with ?? observations.}
\end{abstract}

\section{Introduction}

\begin{figure}[t]
    \centering
    \includegraphics[width=0.9\columnwidth]{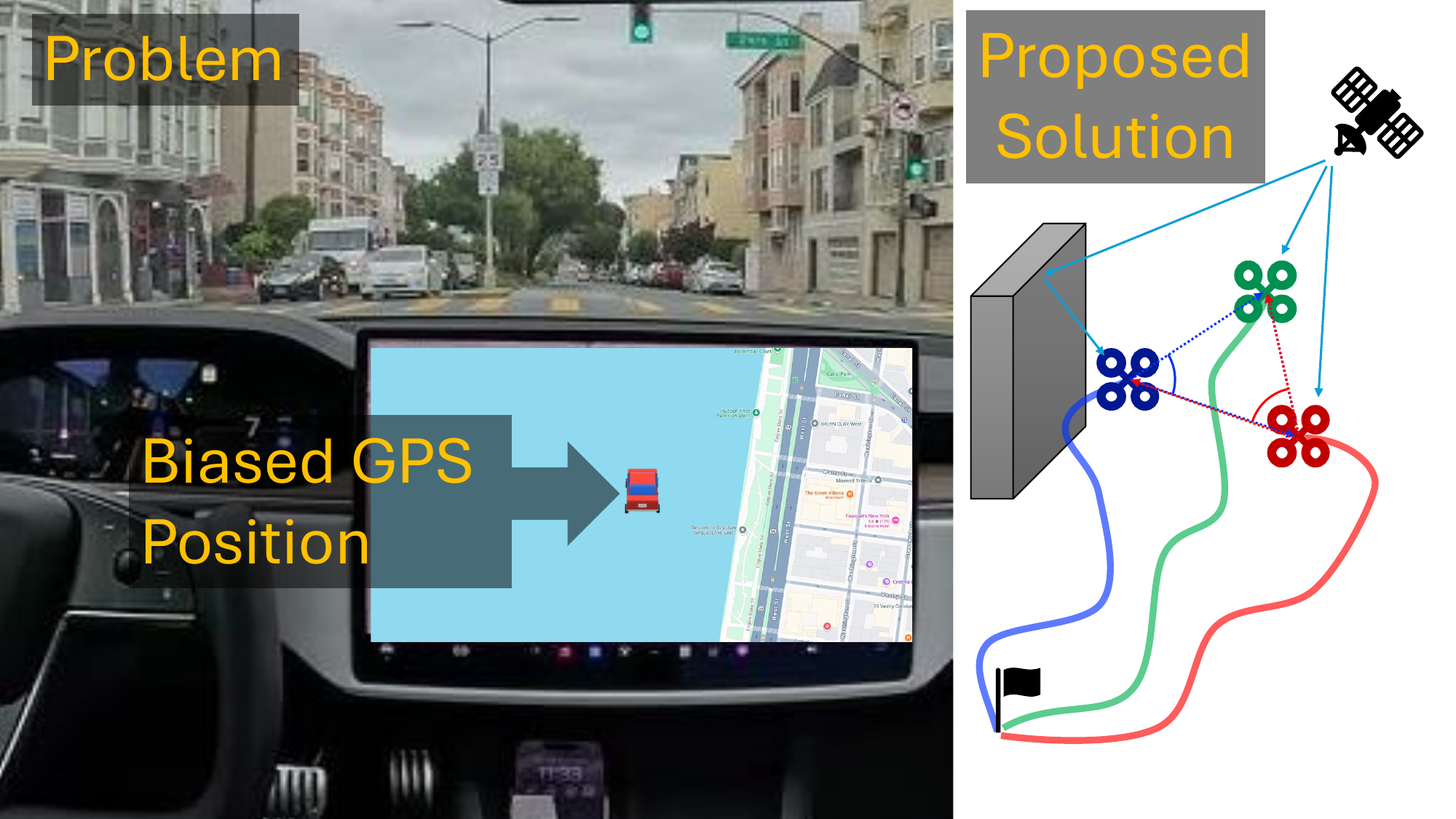}
    \caption{GPS errors during navigation are frustrating, leading to failures of sensor registration and path planning (left).  In this paper we present a method using a team of drones to collaboratively build a map of GPS offsets through collaborative localization and informative path planning (right). See video overview at \href{https://youtu.be/C6F93TBTn8w}{https://youtu.be/C6F93TBTn8w}.
    \label{fig:overview}}
    \vspace*{-0.2in}
\end{figure}

Global Positioning System (GPS) technology is an important tool for localization and navigation in robotics and autonomous systems. However, GPS accuracy can be degraded by atmospheric interference~\cite{intro_atmosphere}, multi-path effects, in which signals reflect off surfaces before reaching the receiver~\cite{intro_multipath}, and occlusions from buildings, trees, or other obstacles in urban and forest environments~\cite{intro_occlusions}. The impacts of these inaccuracies often depend on the environmental context, resulting in localized errors up to several meters~\cite{into_errors}.

For autonomous systems such as unmanned aerial vehicles (UAVs or drones), a small change in positional accuracy can lead to suboptimal decisions or mission failure~\cite{uav_failures}. A drone navigating through a cluttered environment might fail to avoid obstacles or maintain formation with other drones if its localization is unreliable. These limitations require  analysis techniques to address GPS inaccuracies in real time.

This paper uses swarm robotics to address that challenge. Swarm robotics refers to a collective of autonomous systems that work together to achieve a shared objective. By coordinating their actions and sharing information, these systems can exceed the capabilities of individual robots working in isolation. This allows the swarm to efficiently collect and process data for a variety of applications, including environmental monitoring. Localization  is critical for effective coordination and goal-directed navigation.

This work develops methods to estimate static, spatial variations in GPS inaccuracies using collective relative observations (e.g., range and bearing measurements). 
  The majority of GPS inaccuracies are time-varying, but this paper offers an efficient method to characterize static GPS biases. Swarms may be able to apply these methods to dynamic GPS biases.
We define a function that captures how positional errors evolve from one location to another, allowing us to quantify changes in environmental bias.
The data is then used to construct a probabilistic model of GPS errors across the environment.
Using this learned model, drones can dynamically adjust to compensate for environmental inaccuracies, thereby improving overall reliability in challenging conditions.


We developed and integrated a three-part framework for estimating spatial errors in GPS inaccuracies. First, a State Bias Estimation (SBE) algorithm uses a data structure called \emph{deltas} to 
capture the deviation between GPS readings and true positions, as inferred from the relative range of the swarm and bearing measurements to each other. We use a quadratic optimizer to solve for the estimated biases.
Next, the estimated biases at the sampled positions are used to train a Gaussian Process Regression (GPR) model that learns the GPS error function. This model predicts error values at unobserved locations and provides an uncertainty measure that guides where additional data is most valuable.
Finally, an Informative Path Planning (IPP) algorithm leverages the GPR’s uncertainty model, directing drones to the most informative locations~\cite{rw_sgp_Jakkala}. 
%
%
 We validate the proposed methods in a simulated environment. We evaluate the SBE algorithm, including how the process noise affects its performance, and illustrate how the IPP method guides the data collection process.
The effectiveness of the overall process is evaluated using an root mean square error (RMSE) score.

The rest of the paper is organized as follows: Section~\ref{s2} reviews related work. Section~\ref{s3} presents the problem formulation, while Section~\ref{s4} describes the proposed algorithm. Section~\ref{s5} provides the experimental validation, and Section~\ref{s6} concludes the paper.

\section{Related Work} \label{s2}


We draw inspiration from several existing lines of work.

\subsection{Robot Localization} 

 Localization in robotics can be passive or active. \emph{Passive localization}~\cite{rw_passive1,rw_passive2} relies solely on measurements (e.g., GPS signals) or externally selected commands without considering robot motions intended to produce better pose estimates.  \emph{Active localization}~\cite{rw_active1_slam, rw_active2} explicitly plans or adapts the robot’s trajectory to refine the localization process. 

Classical localization approaches often leverage filter-based methods, including the venerable Kalman Filter~\cite{rw_kalman} for linear-Gaussian systems. Since real-world scenarios frequently involve non-linear sensor and motion models, the Extended Kalman Filter (EKF)~\cite{rw_ekf, rw_coop1} linearizes these models via first-order Taylor expansions. Despite its popularity, the EKF can be unreliable if non-linearities are severe or if initial estimates are poor~\cite{rw_coop2}. 

However, because multi-robot systems must frequently fuse both independent and correlated sensor measurements, classical filters can over- or under-confidently weight information when certain correlation statistics are not fully known. To address this challenge, the work in~\cite{rw_coop3} uses the \emph{Split Covariance Intersection Filter} (SCIF), which provides a robust mechanism to handle both known independent components and unknown correlations within the data. Wanasinghe \emph{et al.}~\cite{rw_coop4} use an EKF estimate a robot’s position, and employ an SCIF to fuse measurements received from the swarm.


\subsection{Gaussian Process Regression}

GPR can model continuous fields while providing uncertainty estimates~\cite{rw_Rasmussen2006Gaussian}.  GPR has been used to learn temperature distributions~\cite{rw_gp_temp}, radiation fields~\cite{rw_gp_radiation}, terrain elevations~\cite{rw_gp_terrain}, and signal strength maps~\cite{rw_gp_signal}. Its capacity for interpolating sparse observations into a smooth function with well-defined variance estimates makes GPR suitable for sensor-driven mapping tasks.

De Petrillo \emph{et al.}~\cite{rw_gpr_ugv} use the output of the EKF and process it through both a GPR and Belief Space Planning (BSP) to estimate a UAV's location. In~\cite{rw_gpr_magnetic}, the authors generate indoor magnetic field maps via GPR to individually predict each component of the magnetic field and achieve sub-microTesla accuracies in real-world tests. Meanwhile, the work in~\cite{rw_gpr_rrt} utilizes a GP trained on sparse 3D laser range measurements to derive collision probabilities for obstacle-ridden environments. A rapidly exploring random tree-based path planner then relies on these GP-based probabilities to guide a UAV safely through uncertain spaces.

\subsection{Informative Path Planning (IPP)}

Informative Path Planning (IPP)~\cite{rw_ipp} aims to find trajectories that maximize a robot’s knowledge gain in environmental surveys, such as temperature mapping or object distribution. By prioritizing regions with high expected variance, IPP techniques iteratively reduce uncertainty. Greedy methods~\cite{rw_ipp_greedy}, receding horizon strategies~\cite{rw_ipp_receding}, and Bayesian optimization approaches~\cite{rw_ipp_bo} have all been used to identify the next best measurement location.

One challenge with other IPP methods is the computational overhead of large datasets. Sparse Gaussian Process (SGP) approximations~\cite{rw_sgp} mitigate this issue by approximating the full data set through a reduced set of inducing points. We adopt the SGP method proposed by Jakkala and Akella~\cite{rw_sgp_Jakkala}, which leverages gradient-based path optimization to efficiently scale multi-robot IPP in continuous environments while handling practical constraints such as distance budgets and sensor coverage footprints.
\pagestyle{empty}

\section{Problem Statement} \label{s3}

We assume that there exists an unknown vector-valued error function $ M: \mathbb{R}^2 \to \mathbb{R}^2 $ that describes the GPS bias at each position. The primary objective is to create an estimate of the error function, $\widetilde{M}: \mathbb{R}^2 \to \mathbb{R}^2$, at each position within the environment. 
We assume that the GPS bias function $M(x)$ is continuous
and static, and that  the GPS bias at the starting position of one of the drones is known.


The system consists of a team of $n$ robots, each of which moves in a two-dimensional environment. The robots remain level with the $(x,y)$ plane but can rotate about their $z$ axis. The state space for each robot is $X = \mathbb{R}^2 \times [0, 2\pi)$. The state space for the entire swarm is therefore $\mathcal{X} = X \times \dots \times X = X^n$. The pose of drone $ i $ at any given time $ k $ is denoted by the tuple $x_k^{(i)} = (P_k^{(i)}, \theta_k^{(i)}) $, where $ P_k^{(i)} \in \mathbb{R}^2 $ and $\theta_k^{(i)} \in [0, 2\pi).$ Though the pose $x_k^{(i)}$ is unknown to the robot, each maintains an
estimate $\widehat{x}_k^{(i)}$ of its position based on its prior actions.

The action space $U = [0,s_{\textrm{max}}]\times[\text{-} \omega_{\textrm{max}},\omega_{\textrm{max}}]$ for each drone is defined by $(s_k^{(i)}, \omega_k^{(i)})$, where $s_k^{(i)} \in [0, s_{\text{max}}]$ represents the horizontal speed, and $ \omega_k^{(i)} \in [\text{-}\omega_{\textrm{max}},\omega_{\textrm{max}}] $ represents the angular velocity.

The transition matrix depends on the action $u$ and an additional term $e_k^{(i)}$ accounts for noise introduced during transition. To model $e_k^{(i)}$, we use a small random Gaussian noise vector, where $e_k^{(i)} = (e_x, e_y, 0)^\top$ and $e_x, e_y \sim \mathcal{N}(0, \SI{0.05}{\meter})$. The time period is $T$.
\begin{align}
x_{k+1}^{(i)} = x_k^{(i)} + \begin{pmatrix}
s_k^{(i)} \cos(\theta_k^{(i)}) \\
s_k^{(i)} \sin(\theta_k^{(i)}) \\
 \omega_k^{(i)}
\end{pmatrix} T + e_k^{(i)}.
\end{align}


Each drone is equipped with sensors that provide information about its position and the relative distance and direction to each drone in the swarm. For drone~$i$ at time $k$, each sensor reading is formulated as:

    \renewcommand{\paragraph}[1]{\smallskip\noindent\textbf{#1}:}
    \paragraph{GPS sensor} The reading is $ \mathbf{g}_k^{(i)} \in \mathbb{R}^2 $, which represents the drone's measured position in 2D space as coordinates $(x, y)$. The GPS reading includes an error that is modeled by the error map $M$ at the drone's true position $ P_k^{(i)} $. To model noise, we use a small random Gaussian noise vector $e_{gk}^{(i)}$, where $e_{gk}^{(i)} = (e_{gx}, e_{gy})^\top$ and $e_{gx}, e_{gy} \sim \mathcal{N}(0, \SI{0.01}{\meter})$. 
    \begin{align}
    \mathbf{g}_k^{(i)} = P_k^{(i)} + M(P_k^{(i)}) + e_{gk}^{(i)}.
    \end{align}

    \paragraph{Bearing Sensor} The bearing reading from robot $i$ to robot $j$ at time $k$ is $ \mathbf{b}_k^{(i,j)} \in \mathbb{R}^2 $, a unit vector from drone $i$ pointing towards drone $j$ in the swarm. To model noise, we use a small random Gaussian noise vector $e_{bk}^{(i,j)}$, where $e_{bk}^{(i,j)} = (e_{bx}, e_{by})$ and $e_{bx}, e_{by} \sim \mathcal{N}(0, \SI{0.01}{\meter})$. 
    \begin{align}
    \mathbf{b}_k^{(i,j)} = \frac{P_k^{(i)} - P_k^{(j)} + e_{bk}^{(i,j)} }{\|P_k^{(i)} - P_k^{(j)}  + e_{bk}^{(i,j)} \|}\,.
    \end{align}
    The full bearing measurements made by robot $i$ at time $k$ is $ 
      \mathbf{B}_k^{(i)} = (\mathbf{b}_k^{(i,1)},\, ...\,,
    \mathbf{b}_k^{(i,i-1)},\, \mathbf{b}_k^{(i,i+1)},\, ...\, ,\mathbf{b}_k^{(i,n)}) $,
    a list of $n-1$ bearing measurements to all other drones in the swarm.

    \paragraph{Range Sensor} The reading is denoted by $ r_k^{(i,j)} \in \mathbb{R} $, representing the measured distance from drone $i$ to drone $j$ in the swarm, that is the magnitude of the relative position vector between the two drones. To model noise, we use a small random Gaussian noise vector $e_{rk}^{(i,j)}$, where $e_{rk}^{(i,j)} = (e_{rx}, e_{ry})$ and $e_{rx}, e_{ry} \sim \mathcal{N}(0,\SI{0.01}{\meter})$. The range reading from robot $i$ to robot $j$ at time $k$ is
    \begin{align}
    r_k^{(i,j)} =  ||P_k^{(i)} - P_k^{(j)} + e_{rk}^{(i,j)} || \, .
    \end{align}
    Then the range measurements made by robot $i$ at time $k$ is 
    $
    \mathbf{R}_k^{(i)} = (r_k^{(i,1)},\, ...\,, r_k^{(i,i-1)},\, r_k^{(i,i+1)},\, ...\,,r_k^{(i,n)})
    $,
    a list of $n-1$ range measurements to all other drones in the swarm.

\bigskip

The observation space is $ \mathcal{Y} = \mathbb{R}^2\times \mathbb{R}^{2(n-1)} \times \mathbb{R}^{n-1}$, and the observation function is 
\begin{align}
y_k^{(i)} = 
\begin{pmatrix}
\mathbf{g}_k^{(i)} \\
\mathbf{B}_k^{(i)} \\
\mathbf{R}_k^{(i)}
\end{pmatrix}  .
\end{align}
%
The goal is to learn an estimate of the error function
\begin{align}
\widetilde{M}: \mathbb{R}^2 \to \mathbb{R}^2,
\end{align}
which captures spatially varying GPS bias at each position. 
%
%
Two forms of input can provide data to estimate $\widetilde{M}$:
\begin{enumerate}
    \item \textbf{Passive Version}: We assume each drone $i$ has already performed a sequence of actions $u_1^{(i)},\dots, u_K^{(i)}$ and logged corresponding observations $y_1^{(i)},\dots,y_K^{(i)}$. 
    \item \textbf{Active Version}: We intentionally choose new actions $u_1^{(i)},\dots, u_K^{(i)}$ that move drones to unvisited areas to gather additional observations $y_1^{(i)},\dots,y_K^{(i)}$. 
\end{enumerate}

In both cases, we evaluate performance using the root mean squared error (RMSE) between the error function $M$ and the estimated error function $\widetilde{M}$.

\section{Algorithm Description} \label{s4}
This section outlines the process for estimating the spatial distribution of GPS errors, starting with the \emph{State Bias Estimation} (SBE) algorithm, which estimates the spatial bias at each estimated position (Sec.~\ref{subsec:SBE}). Next, these refined data points serve as input to a GPR model, building a continuous error map that predicts the GPS bias at unobserved locations across the environment (Sec.~\ref{subsec:GPRestimate}).
For the active version, we integrate an IPP method which leverages the uncertainty in this error map to select high-value sampling locations, directing the drones to areas where additional measurements can most effectively improve the error estimation (Sec.~\ref{subsec:IPP}). Through this iterative loop of bias correction, error map updates, and targeted exploration, the swarm continually refines its understanding of the GPS inaccuracies.

\subsection{State Bias Estimation (SBE) Algorithm}\label{subsec:SBE}

The SBE algorithm is designed to estimate the biases at the estimated positions based on collected data by the UAVs.
%
A \emph{delta} is a 3-tuple $(\mathbf{p}_{1}, \mathbf{p}_{2}, \delta)$ in which $\mathbf{p}_{1},\mathbf{p}_{2},\delta \in \mathbb{R}^3$. The idea is that, between positions $\mathbf{p}_{1}$ and $\mathbf{p}_{2}$, the GPS bias changes by $\delta$. This concept is motivated by the fact that, though the robots can observe relative changes in the GPS measurements, they cannot make absolute measurement of GPS bias.

The SBE algorithm is takes as input a set of $m$ deltas
\begin{align}
\Delta = \{ (\mathbf{p}_{1}^{(1)}, \mathbf{p}_{2}^{(1)}, \delta^{(1)}),  \ldots, (\mathbf{p}_{1}^{(m)}, \mathbf{p}_{2}^{(m)}, \delta^{(m)}) \},
\end{align}
and produces as output an estimate of the bias at each of the $2m$ positions mentioned in $\Delta$.

To accomplish this, we solve the following quadratic optimization problem~\cite{qp_solver}:

\paragraph{Variables} We use two scalar variables $b_x^{(i)},b_y^{(i)}$ to represent the bias at each position in $\Delta$. There may be less than $4m$ variables if a position appears in more than one delta.
    For a given delta $(\mathbf{p}_1^{(i)},\mathbf{p}_2^{(i)},\delta^{(i)})$ we write $b_x(\mathbf{p}_1^{(i)})$, $b_y(\mathbf{p}_1^{(i)})$ for the bias variables associated with position $\mathbf{p}_1^{(i)}$ and likewise for $\mathbf{p}_2^{(i)}$.
    
    \paragraph{Constraints} To remove the global ambiguity, we add the constraint $(b_x^{(1)},b_y^{(1)} ) = ( b_x^{(\textrm{init})},b_y^{(\textrm{init})} )$ where $b_x^{(1)},b_y^{(1)}$ is the bias of some designated reference position and $b_x^{(\textrm{init})},b_y^{(\textrm{init})}$ is its known bias vector. This condition anchors all other biases so that they are expressed relative to $b_x^{(1)},b_y^{(1)}$.
    
    \paragraph{Objective function} We introduce a least-squares criterion that sums the squared residuals across all deltas: 
%
\begin{align}
C = \sum_{(\mathbf{p}_a,\mathbf{p}_b,\delta)\in \Delta} \Bigl[
    &\bigl(b_x(\mathbf{p}_a) - b_x(\mathbf{p}_b) - \delta_{x}\bigr)^2  \nonumber \\ + & \bigl(b_y(\mathbf{p}_a) - b_y(\mathbf{p}_b) -\delta_{y}\bigr)^2 \Bigr].
\end{align}

Minimizing $C$ over all unknown biases yields the set of biases that best fits the observed deltas.

Thus far, we have described bias estimation without discussing how the swarm can obtain deltas. This section describes how our system acquires deltas. In both cases, we produce a \(\delta\) that captures the discrepancy in GPS readings relative to either inter-drone comparisons or single-drone movements.
We assume that the robots estimate their own positions as they move based on dead reckoning. We represent this as $\hat{p}_k^{(i)} \in \mathbb{R}^2$, considering only the spatial component.
We form deltas in two distinct ways, as shown in Fig.~\ref{fig:deltas}:
\definecolor{myLightBlue}{RGB}{141,141,248} 
\definecolor{myDarkGreen}{RGB}{48, 111 ,29}

    
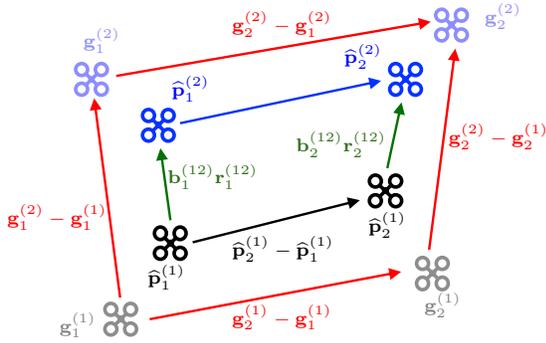
\begin{figure}[t]
    \centering
    \begin{overpic}[width=0.3\textwidth]{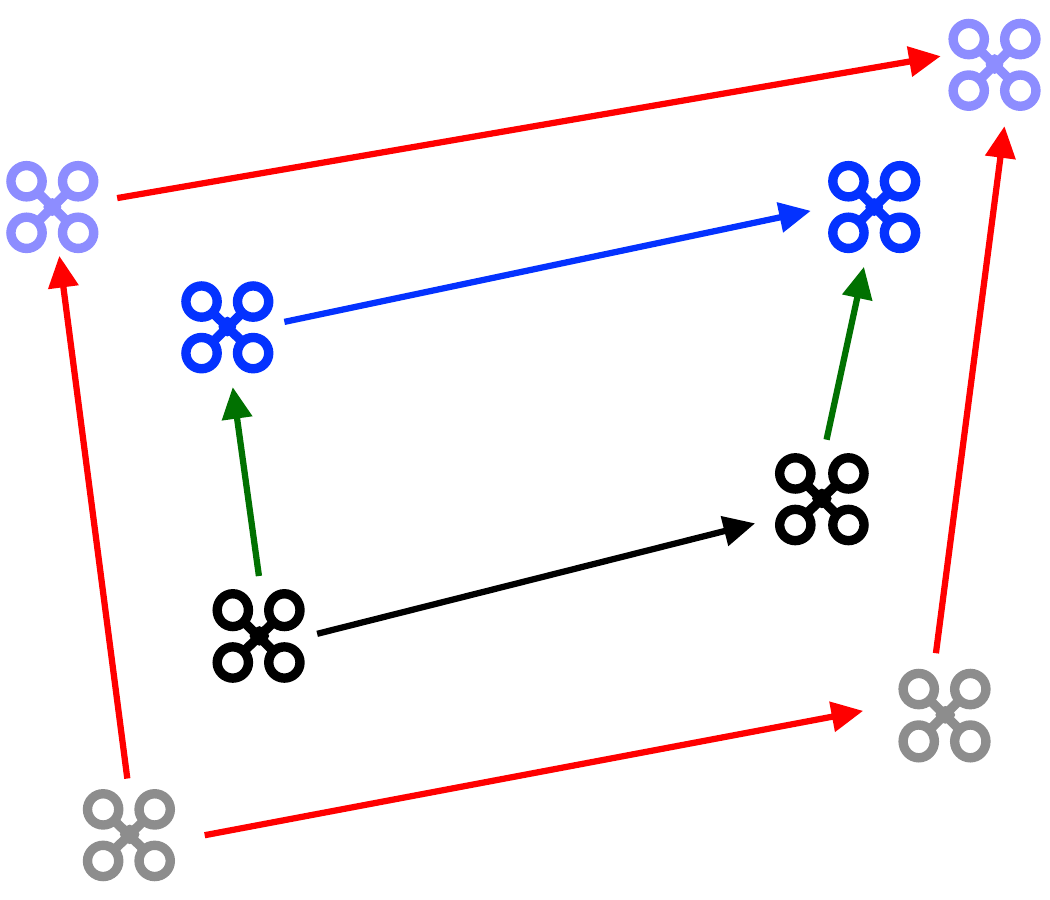}
      \put(-3,3){\color{gray}\scriptsize $\mathbf{g}_1^{(1)}$}
      \put(88,8){\color{gray}\scriptsize $\mathbf{g}_2^{(1)}$}
      \put(-16,30){\color{red}\scriptsize $\mathbf{g}_1^{(2)}- \mathbf{g}_1^{(1)}$}
      \put(94,50){\color{red}\scriptsize $\mathbf{g}_2^{(2)}- \mathbf{g}_2^{(1)}$}
      \put(40,5){\color{red}\scriptsize $\mathbf{g}_2^{(1)}- \mathbf{g}_1^{(1)}$}
      \put(3,74){\color{myLightBlue}\scriptsize $\mathbf{g}_1^{(2)}$}
      \put(103,80){\color{myLightBlue}\scriptsize $\mathbf{g}_2^{(2)}$}
      \put(40,78){\color{red}\scriptsize $\mathbf{g}_2^{(2)}- \mathbf{g}_1^{(2)}$}
      \put(19,15){\color{black}\scriptsize $\widehat{\mathbf{p}}_1^{(1)}$}
      \put(74,28){\color{black}\scriptsize $\widehat{\mathbf{p}}_2^{(1)}$}

      \put(25,62){\color{blue}\scriptsize $\widehat{\mathbf{p}}_1^{(2)}$}
      \put(68,70){\color{blue}\scriptsize $\widehat{\mathbf{p}}_2^{(2)}$}
      
      \put(40,22){\color{black}\scriptsize $\widehat{\mathbf{p}}_2^{(1)}-\widehat{\mathbf{p}}_1^{(1)}$}
      \put(24,40){\color{myDarkGreen}\scriptsize $\mathbf{b}_1^{(12)}\mathbf{r}_1^{(12)}$}
      \put(56,48){\color{myDarkGreen}\scriptsize $\mathbf{b}_2^{(12)}\mathbf{r}_2^{(12)}$}
    \end{overpic}
    \caption{A simple two-drone scenario collecting \emph{delta} measurements.
    Solid color icons show estimated drone positions ($\widehat{\mathbf{p}}_1^{(1)}$), while lighter icons show GPS measurements ({\color{gray}$\mathbf{g}_1^{(1)}$}).
    }
    \label{fig:deltas}
\end{figure}

    \paragraph{Two Drones (Same Time)} At each time step $k$ and for each pair of distinct drones $i$ and $j$, the discrepancy between GPS readings of drone~$i$ and drone~$j$ leads to a delta of the form
\begin{align}
    \bigl(\mathbf{g}_k^{(i)},\, \mathbf{g}_k^{(j)},\, \mathbf{g}_k^{(i)} - \mathbf{g}_k^{(j)} - r_k^{(i,j)} \mathbf{b}_k^{(i,j)} \bigr) \, .
    \end{align}

    \paragraph{One Drone (Consecutive Timesteps)} When a single drone moves from time $k-1$ to time $k$, we form a delta from the estimated positions of the form 
    \begin{align}
    \bigl(\mathbf{g}_k^{(i)},\mathbf{g}_{k-1}^{(i)},(\mathbf{g}_k^{(i)} - \hat{\mathbf{p}}_k^{(i)}\bigr) -
    \bigl(\mathbf{g}_{k-1}^{(i)} - \hat{\mathbf{p}}_{k-1}^{(i)})\bigr) \, .
    \end{align}

With $n$ drones and $k$ time steps, we collect $m = kn^2$ deltas. Each drone per step collects $n$ deltas where $(n-1)$ deltas are to the other drones in the swarm, and one delta between the drone's time steps.

\subsection{Estimating the GPS Error Function with GPR}\label{subsec:GPRestimate}
To calculate $\widetilde{M}$, we model the GPS bias vector at a spatial position using Gaussian Process Regression (GPR). The training data consists of estimated positions and corresponding bias vectors returned by the SBE.
The GPR is a non-parametric, Bayesian machine learning approach that is particularly well-suited for modeling spatial data \cite{rw_Rasmussen2006Gaussian, gp_intro1}. A Gaussian Process (GP) defines a distribution over functions, and its predictions are characterized by a mean function $\mu(\mathbf{x})$ and a covariance function $k(\mathbf{x}, \mathbf{x}')$.
%
%
We use the Radial Basis Function (RBF) kernel \cite{gp_kernel}, which ensures smooth interpolation between sampled points while maintaining a well-defined idea of uncertainty in unexplored areas. The resulting predictive mean $\mu(\mathbf{x})$ represents the estimated error vector field, while the predictive variance $\Sigma(\mathbf{x})$ quantifies estimation uncertainty.

\subsection{Informative Path Planning}\label{subsec:IPP}
Our informative path planning method has two parts.

\subsubsection{Sparse Gaussian Process for Inducing Placements}
Using a grid of evenly sampled points that cover the entire environment, we use the variance function returned by the GPR to identify a set of $p$ placements per drone for each of the $n$ drones that reduce uncertainty in the GP model. Hence, $np$ total points will be selected in $\mathbb{R}^2$.
Using the SGP approach \cite{rw_sgp_Jakkala}, we solve an optimization problem over an initial guess of inducing points, adjusting them to minimize the overall variance in the learned map. 

\subsubsection{Drone Route Assignment}
Once we obtain $n$ distinct routes, we assign them to the $n$ drones to minimize total travel time to the initial point in the route. Let $\{\mathcal{R}_1,\dots,\mathcal{R}_n\}$ be the computed routes, and let $\{P_1,\dots,P_n\}$ be the current positions of the drones. Following \cite{burkard_assignment_problem}, we define a cost matrix $C \in \mathbb{R}^{n\times n}$ by
\begin{align}
C_{ij} = \bigl\|\, P_i - \mathcal{R}_j[0] \bigr\| \, ,
\end{align}
which is the Euclidean distance from the $i$\textsuperscript{th} drone’s current location to the first point in route $j$. Solving a linear assignment problem on $C$ (using SciPy 1.0)
yields an optimal pairing between drones and routes 
\begin{align}
\min_{\pi}\quad \sum_{i=1}^n \bigl\|\, P_i - \mathcal{R}_{\pi(i)}[0] \bigr\| \, ,
\end{align}
where $\pi$ is a permutation of $\{1,\dots,n\}$. Once matched, each drone $i$ is assigned to the route $\mathcal{R}_{\pi(i)}$, thereby minimizing total travel distance.

With the routes assigned, each drone follows its designated path $\mathcal{R}_i$, collecting new measurements. Once the designated paths are completed by all the drones, then each drone resamples a new path based on the updated variance function.

\section{Experimental Validation} \label{s5}
\begin{figure}[t]
    \centering
    \includegraphics[width=1\linewidth]{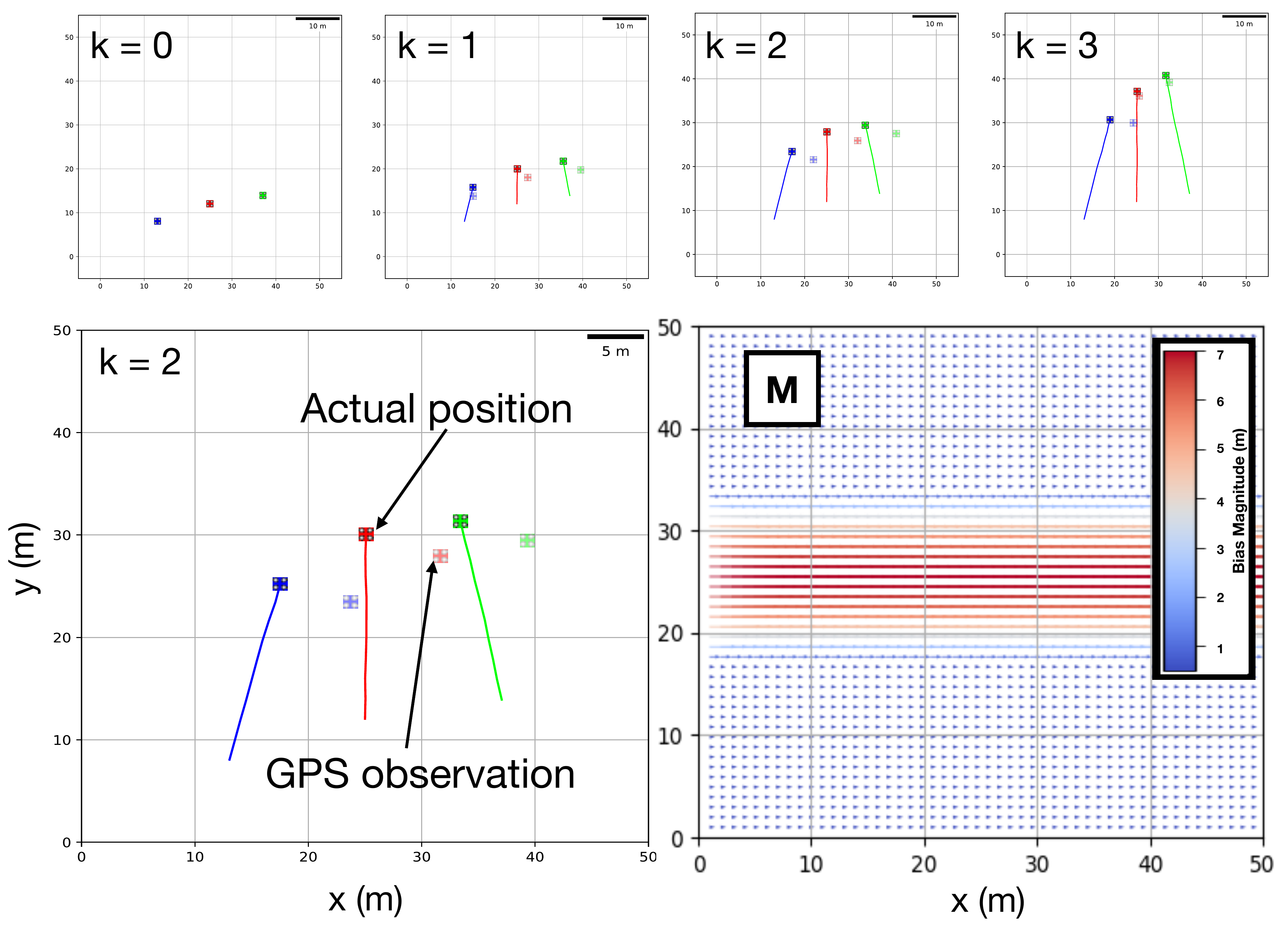}%
    \caption{Bias map influence $M$ (lower right) on three simulated drones at four time steps. The GPS reading from the previous time step is shown along with the actual drone position.
    \todo{Fig 3 is supposed to show the affect of the bias map on the GPS readings.  The bias map is a shift in +x, but the GPS readings compared to the actual robot locations is a shift in both x and y -- how can this be?  Should there be a path for the GPS measurements - I guess that it can be modified to include the end of the experiment in the big trajectory plot}}
    \label{fig:sim_fig_2}
\end{figure}


In this section, we present an experimental validation of our proposed framework, which integrates the SBE, GPR, and IPP methods to produce the estimated bias map $\widetilde{M}$. We begin by validating the SBE output and then assess how effectively the GPR reconstructs an accurate bias map from that data in a simplified two-dimensional environment. Then, we demonstrate how these components, coupled with the SGP, can generate a map of GPS biases.

All experiments were conducted in a custom simulator designed to simulate multi-drone scenarios and collect the necessary data for our study. Our simulation environment with three representative drones is shown in Fig.~\ref{fig:sim_fig_2}. The top shows four time frames of the drones moving to a goal location. The figure illustrates how drones traverse the environment while demonstrating the impact of the bias map on their GPS measurements.

\subsection{Validating the SBE}

\begin{figure}[t]
    \centering
    \includegraphics[width=0.5\textwidth]{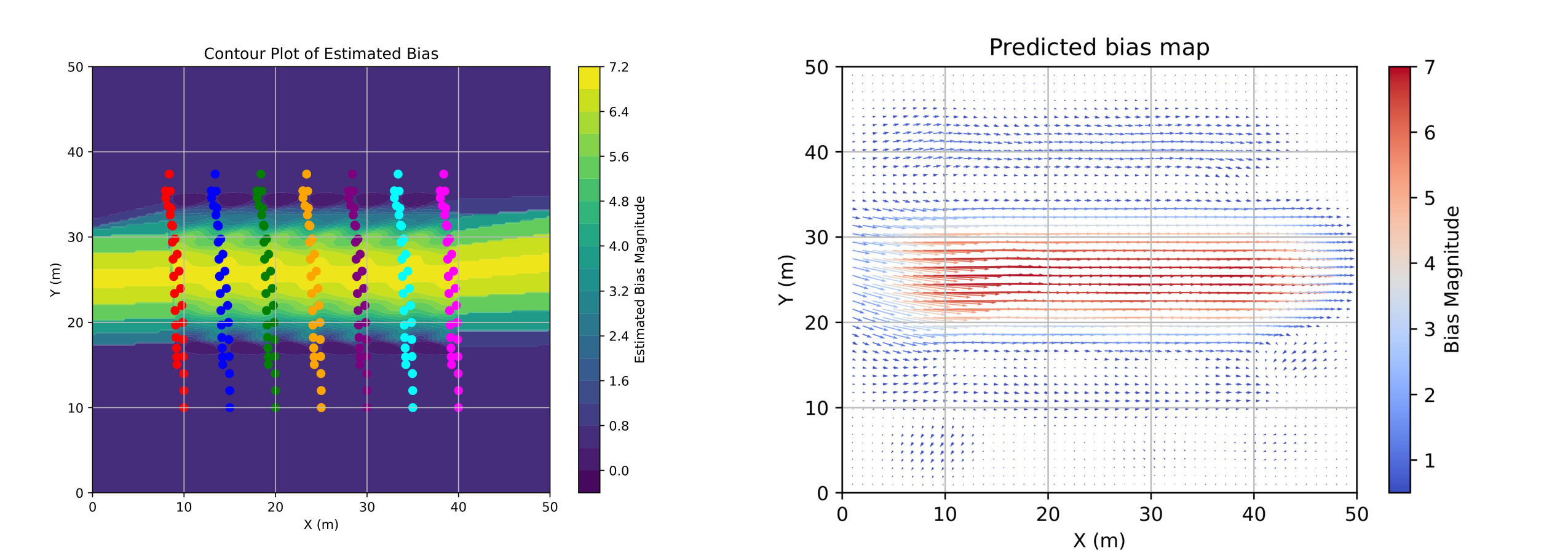}
    \vspace{-1em}
    \caption{left: Difference in the True Biases $M$ vs Estimated Biases $\widetilde{M}$ for the 210 positions visited by seven drones with $\SI{0.05}{\meter}$ process noise. Right: Predicted bias map $\widetilde{M}$ after 7 drones collected 210 total measurements (vectors shown are scaled with their length). 
    }
    \label{fig:solver_diff}
\end{figure}

To test the SBE, we use a $\SI{50}{\meter}\times \SI{50}{\meter}$ field map with a known bias function shown in Fig.~\ref{fig:sim_fig_2} on the bottom right. We used $n=7$, with drones that moved between two designated waypoints. We assumed that drone one's initial bias is known at $k=0$, which is used as the anchor point in the SBE. We conducted 10 trials that varied the standard deviation of the process noise from $\SI{0}{\meter}$ to $\SI{0.5}{\meter}$ in increments of $\SI{0.05}{\meter}$ to be consistent with those modeled in prior work. The drones each collected data for a fixed duration, with process noise standard deviation of $\SI{0.05}{\meter}$, yielding 210 total measured positions shown in Fig.~\ref{fig:solver_diff}(a).
 We computed deltas from collected positional data and passed them to the SBE solver. We then trained a GPR model on the SBE outputs, obtaining a bias map over the domain.

\begin{figure}[tbp]
    \centering
\includegraphics[width=0.4\textwidth]{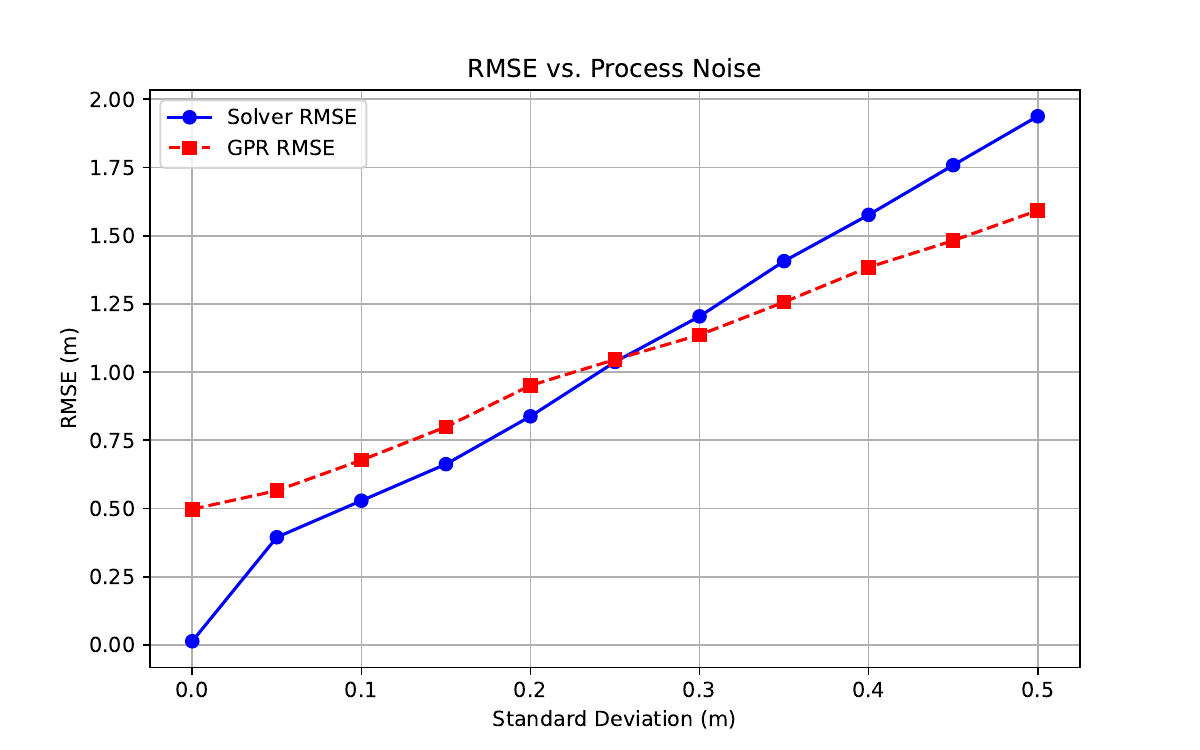}
\vspace{-1em}
    \caption{Effect on RMSE based on varying process noise levels. The Solver RMSE is only calculated over the points sampled, while the GPR RMSE is calculated over a 2D grid with 1 meter spacing over the workspace.
    \label{fig:RMSE_noise}}
\end{figure}

We evaluate the effectiveness of this process on
%
    SBE RMSE (the root mean squared error of the solver’s estimated biases against the known bias function evaluated at those points) and
    GPR RMSE (the root mean squared error of the final estimated bias function against the known bias function across the entire environment.
%
%
%
The results slow that the solver's output RMSE is proportional to the process noise. This behavior is shown in Fig.~\ref{fig:RMSE_noise} for process noise standard deviation ranging from $\SI{0}{\meter}$ to  $\SI{0.5}{\meter}$. 
After passing the solver outputs to the GPR, the estimated bias map yields a final RMSE of $\SI{1.06}{\meter}$ relative to the field’s known bias function. Despite the noise, the GPR correctly recovers the general trend of the biases, as illustrated in Fig.~\ref{fig:solver_diff}(b) compared to $M$ in Fig.~\ref{fig:sim_fig_2} on the bottom right. 

These results indicate that the SBE reconstructs the environment’s bias field in Fig.~\ref{fig:RMSE_noise}. In particular, the estimated bias vectors generally agree with the true bias vectors as seen in the difference visualization, confirming that using pairs of positions, as proposed above, provides enough information to accurately capture the bias. 



\subsection{Validating the IPP}

\begin{figure}[t]
    \centering
    \includegraphics[width=1\columnwidth]{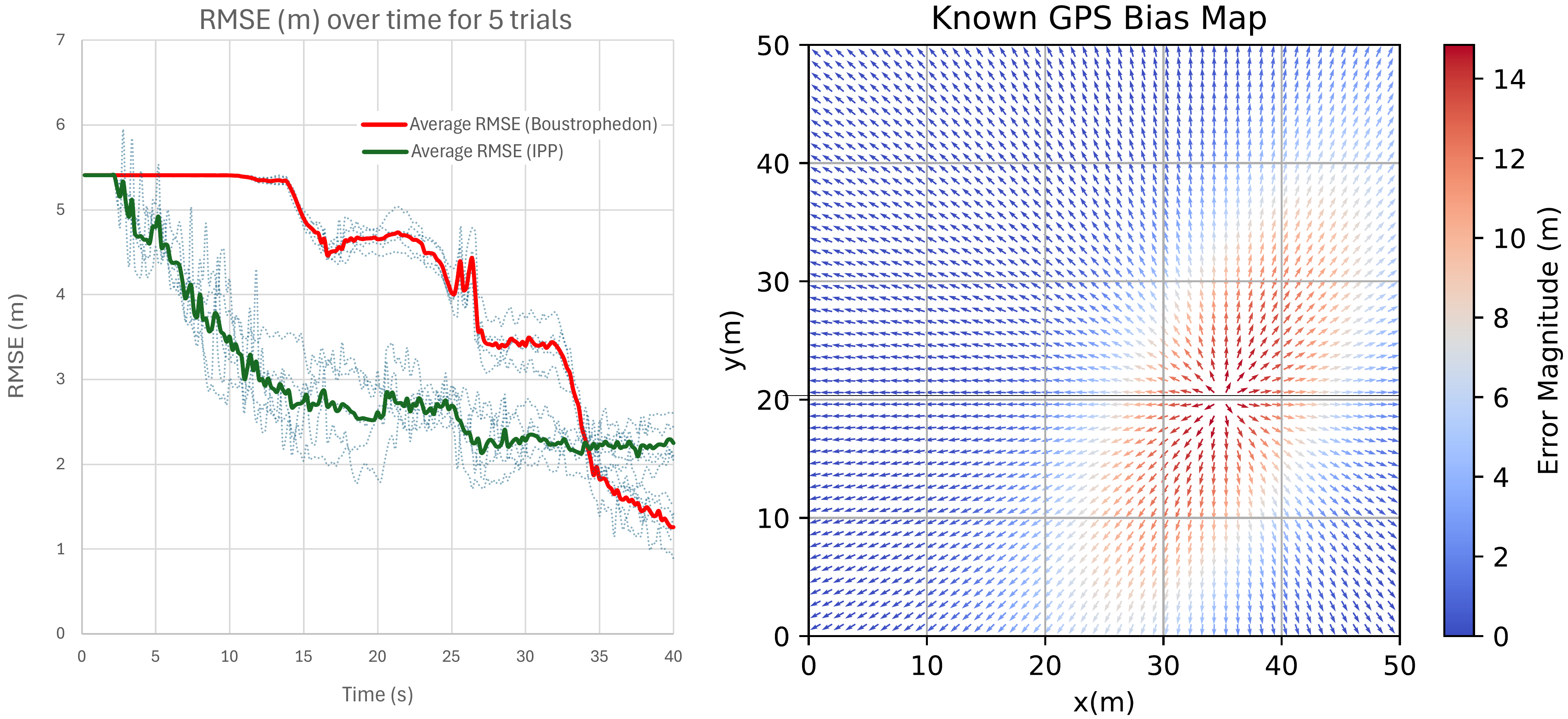}
    \vspace{-1em}
    \caption{
        (left) Comparison of RMSE calculated by IPP and Boustrophedon paths for three drones over five trials for each method. The averages are shown with thick lines.  (right) The error function $M$ to be estimated.
    }
    \label{fig:knownM_and_RMSE}
\end{figure}

We demonstrate the application of the SGP approach in a two-dimensional field measuring 
$\SI{50}{\meter}\times \SI{50}{\meter}$. This experiment used an error map featuring a Gaussian-magnitude vector field centered at $(\SI{35}{\meter}, \SI{30}{\meter})$ and pointing radially outwards shown in Fig.~\ref{fig:knownM_and_RMSE} on the right. 

To evaluate the effectiveness of the IPP approach, a performance comparison against a standard open-loop method is presented.
This experiment compares the performance of two path planning methods, boustrophedon (back-and-forth seed sowing path), and IPP.  Both use GPR to build the bias map, and are compared using the RMSE of the GPR-generated bias map as the system runs.
 In this case, we used $n=3$ drones. The IPP uses $p=3$ placement points for this experiment at each planning step, recalculating after the final point is reached. 
 While IPP adaptively selects sampling locations to improve bias estimation, the open-loop approach follows a predefined trajectory without dynamically adjusting based on collected data.

Both methods were evaluated under identical initial conditions. The experiments were conducted in the same $\SI{50}{\meter}\times \SI{50}{\meter}$ field with the bias map. 
We report the RMSE of the map five times a second.  The state bias algorithm started at two seconds. Before two seconds, an all-zero bias map estimate was used.

\begin{figure}[t]
    \centering
\includegraphics[width=\linewidth]{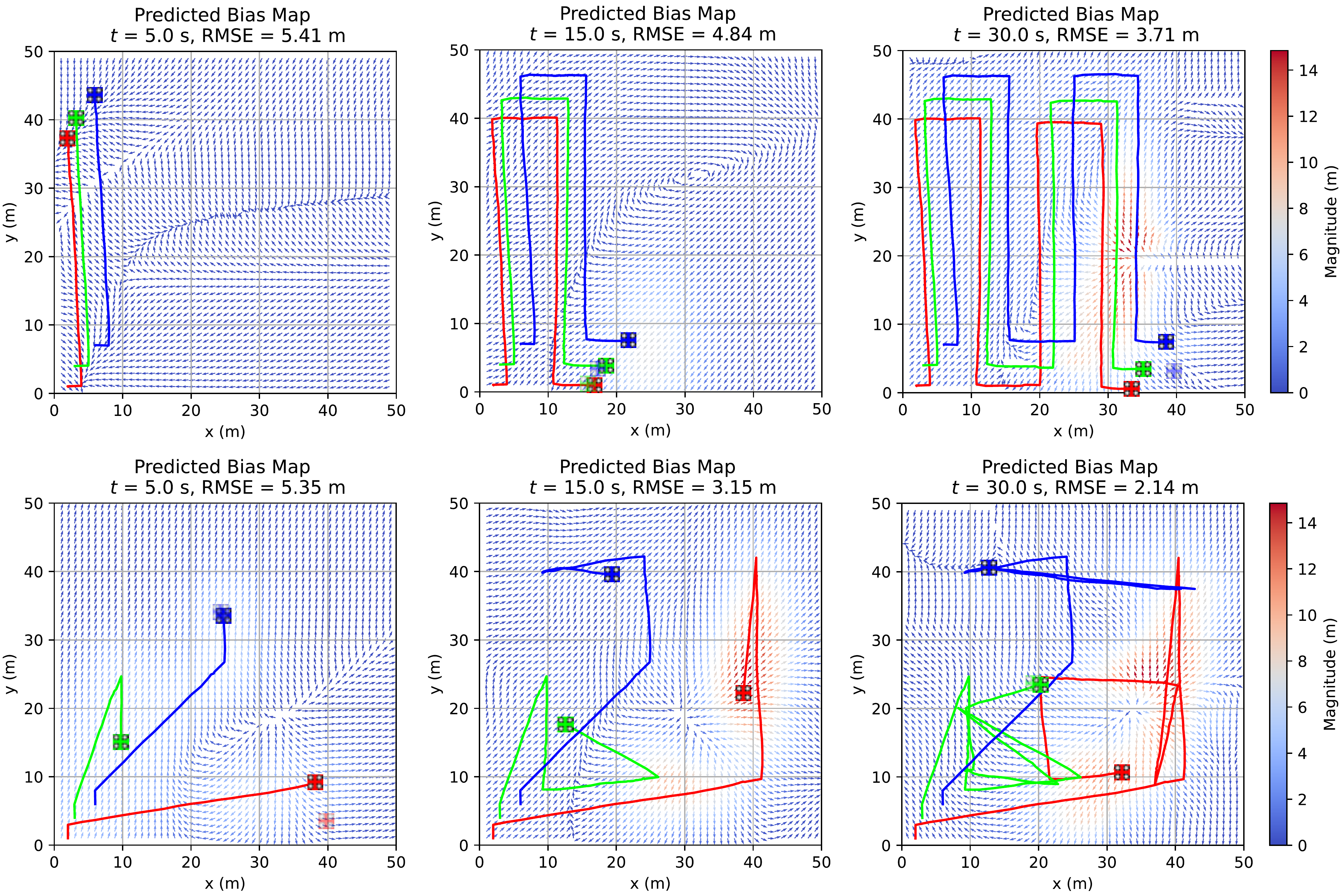}
    \caption{
        Frames showing drone positions and evolving bias map with Boustrophedon coverage on top row and IPP coverage on bottom row. See side-by-side comparison in video attachment \href{https://youtu.be/C6F93TBTn8w}{https://youtu.be/C6F93TBTn8w}.
    }
    \label{fig:CompareBoustrophedonIPPMaps}
\end{figure}

The selected bias maps generated by the Boustrophedon and IPP system are shown in Fig.~\ref{fig:CompareBoustrophedonIPPMaps}, which show frames at times $[5, 15, 30]$ seconds. 
%
The RMSE is reported on the left side of Fig.~\ref{fig:knownM_and_RMSE} over five trials for each method. As expected, the IPP initially converges faster (in less than five seconds) to the map than the open-loop method.

\section{Conclusion} \label{s6}

Our work proposes a framework for estimating spatial variations in a static GPS bias and demonstrates the effectiveness of that framework. The SBE reliably reconstructs the biases from a set of deltas, and the GPR is able to generate a coherent error map from the SBE outputs. In addition, our integration of IPP further refines the bias map by guiding the drones to collect data in high-uncertainty regions. While process and measurement noise introduce some deviations in the estimated biases, the overall performance remains reasonable, confirming that our method can capture the environmental bias. The IPP prioritizes informative exploration, reducing the number of required measurements and completing the process in less time. However, this comes at the cost of a slightly higher RMSE compared to an open-loop strategy, which achieves a more accurate estimation with a more controlled exploration strategy and less movement. This suggests a possible trade-off between accuracy and efficiency in environmental coverage.


\bibliographystyle{IEEEtran}
\bibliography{IEEEabrv,biblio.bib}

\end{document}